\documentclass[letterpaper]{article} %
\usepackage[]{aaai24}  %
\usepackage{times}  %
\usepackage{helvet}  %
\usepackage{courier}  %
\usepackage[hyphens]{url}  %
\usepackage{graphicx} %
\usepackage{natbib}  %
\usepackage{caption} %
\usepackage{algorithm}
\usepackage{algorithmic}

\usepackage{newfloat}
\usepackage{listings}
\DeclareCaptionStyle{ruled}{labelfont=normalfont,labelsep=colon,strut=off} %
\floatstyle{ruled}
\newfloat{listing}{tb}{lst}{}
\floatname{listing}{Listing}
\usepackage{amsfonts,amsmath,amssymb,amsthm,bm}
\usepackage{cleveref}
\usepackage{booktabs}
\usepackage{enumitem}
\usepackage{xcolor}
\usepackage{times}

\newtheorem{theorem}{Theorem}
\newtheorem{proposition}[theorem]{Proposition}
\theoremstyle{definition}
\newtheorem{definition}[theorem]{Definition}

\definecolor{Gred}{RGB}{219, 50, 54}
\definecolor{Ggreen}{RGB}{60, 186, 84}
\definecolor{Gblue}{RGB}{72, 133, 237}
\definecolor{Gyellow}{RGB}{247, 178, 16}
\definecolor{ToCgreen}{RGB}{0, 128, 0}
\definecolor{myGold}{RGB}{231,141,20}
\definecolor{myBlue}{rgb}{0.19,0.41,.65}
\definecolor{myPurple}{RGB}{175,0,124}

\providecommand{\Comments}{0}  %
\ifnum\Comments=1
\usepackage[colorinlistoftodos,prependcaption,textsize=scriptsize]{todonotes}
\paperwidth=\dimexpr \paperwidth + 5cm\relax
\oddsidemargin=\dimexpr\oddsidemargin + 2.5cm\relax
\evensidemargin=\dimexpr\evensidemargin + 2.5cm\relax
\marginparwidth=\dimexpr\marginparwidth + 1.6cm\relax
\else
\usepackage[disable]{todonotes}
\fi
\newcommand{\mytodo}[1]{\ifnum\Comments=1{#1}\fi}

\newcommand{\tableoftodos}{\ifnum\Comments=1 \listoftodos[Comments/To Do's] \fi}

\def\ddefloop#1{\ifx\ddefloop#1\else\ddef{#1}\expandafter\ddefloop\fi}
\def\ddef#1{\expandafter\def\csname b#1\endcsname{\ensuremath{{\bm #1}}}}
\ddefloop ABCDEFGHIJKLMNOPQRSTUVWXYZabcdeghijklnopqrstuvwxyz\ddefloop %
\def\ddef#1{\expandafter\def\csname c#1\endcsname{\ensuremath{\mathcal{#1}}}}
\ddefloop ABCDEFGHIJKLMNOPQRSTUVWXYZ\ddefloop
\def\ddef#1{\expandafter\def\csname #1\endcsname{\ensuremath{\mathbb{#1}}}}
\ddefloop ABCDEFGHIJKLMNOPQRSTUVWXYZ\ddefloop
\def\ddef#1{\expandafter\def\csname t#1\endcsname{\ensuremath{\tilde{#1}}}}
\ddefloop abcdefghijklmnpqrstuvwxyz\ddefloop  %
\def\ddef#1{\expandafter\def\csname h#1\endcsname{\ensuremath{\hat{#1}}}}
\ddefloop y\ddefloop  %

\newcommand{\eps}{\varepsilon}
\DeclareMathOperator*{\Ex}{\mathbb{E}}

\newcommand{\DP}{\mathsf{DP}}
\newcommand{\RR}{\mathsf{RR}}
\newcommand{\SGD}{\mathsf{SGD}}
\newcommand{\DPSGD}{\DP\text{-}\SGD}
\newcommand{\Hybrid}{\mathsf{Hybrid}}
\newcommand{\AUC}{\mathrm{AUC}}
\newcommand{\np}{\mathrm{np}}

\newcommand{\sens}{\bullet}
\newcommand{\nonsens}{\circ}
\newcommand{\cXns}{\mathcal{X}^{\nonsens}}
\newcommand{\cXs}{\mathcal{X}^{\sens}}
\newcommand{\xns}{x^{\nonsens}}
\newcommand{\xs}{x^{\sens}}

\newcommand{\txs}{\tilde{x}^{\sens}}
\newcommand{\dns}{d_{\nonsens}}
\newcommand{\ds}{d_{\sens}}
\newcommand{\wns}{\bw^{\nonsens}}
\newcommand{\ws}{\bw^{\sens}}
\newcommand{\wc}{\bw^{\mathrm{c}}}

\title{Training Differentially Private Ad Prediction Models with Semi-Sensitive Features}
\author{
  Lynn Chua\textsuperscript{\rm 1},
  Qiliang Cui\textsuperscript{\rm 1},
  Badih Ghazi\textsuperscript{\rm 1},
  Charlie Harrison\textsuperscript{\rm 1},
  Pritish Kamath\textsuperscript{\rm 1},
  Walid Krichene\textsuperscript{\rm 1},
  Ravi Kumar\textsuperscript{\rm 1},
  Pasin Manurangsi\textsuperscript{\rm 1},
  Krishna Giri Narra\textsuperscript{\rm 1},
  Amer Sinha\textsuperscript{\rm 1},
  Avinash Varadarajan\textsuperscript{\rm 1},
  Chiyuan Zhang\textsuperscript{\rm 1},
}
\affiliations{
  \textsuperscript{\rm 1}Google
}

\begin{document}

\maketitle

\begin{abstract}
Motivated by problems arising in digital advertising, we introduce the task of training differentially private (DP) machine learning models with semi-sensitive features. In this setting, a subset of the features is known to the attacker (and thus need not be protected) while the remaining features as well as the label are unknown to the attacker and should be protected by the DP guarantee. This task interpolates between training the model with \emph{full} DP (where the label and all features should be protected) or with \emph{label} DP (where all the features are considered known, and only the label should be protected). We present a new algorithm for training DP models with semi-sensitive features.  Through an empirical evaluation on real ads datasets, we demonstrate that our algorithm surpasses in utility the baselines of (i) DP stochastic gradient descent (DP-SGD) run on all features (known and unknown), and (ii) a label DP algorithm run only on the known features (while discarding the unknown ones).
\end{abstract}

\section{Introduction}

In recent years, large-scale machine learning (ML) algorithms have been adopted and deployed for different ad modeling tasks, including the training of predicted click-through rates (a.k.a. pCTR) and predicted conversion rates (a.k.a. pCVR) models. Roughly speaking, pCTR models predict the likelihood that an ad shown to a user is clicked, and pCVR models predict the likelihood that an ad clicked (or viewed) by the user leads to a \emph{conversion}---which is defined as a useful action by the user on the advertiser site or app, such as the purchase of the advertised product.

Heightened user expectations around privacy have led different web browsers (including Apple Safari \cite{safari},  Mozilla Firefox \cite{mozilla}, and Google Chrome \cite{chromium}) to the deprecation of third-party cookies, which are cross-site identifiers that had hitherto allowed the joining in the clear of the datasets on which the pCTR and pCVR models are trained. More precisely, third-party cookies previously allowed determining the conversion label for pCVR models as well as the construction of features (for pCTR and pCVR models) that depend on the user’s behavior on sites other than the publisher where the ad was shown.

In order to support essential web functionalities that are affected by the deprecation of third-party cookies, different web browsers have been building privacy-preserving APIs, including for ads measurement and modeling such as the Interoperable Private Attribution (IPA) developed by Mozilla and Meta \cite{ipa-blog},  Masked LARK from Microsoft \cite{pfeiffer2021masked, MaskedLARk-github}, the Attribution Reporting API, available on both the Chrome browser \cite{chrome-attribution-reporting} and the Android operating system \cite{attribution-reporting-api-android}, and the Private Click Measurement (PCM) \cite{pcm-safari} and Private Ad Measurement (PAM) APIs \cite{apple-prio-like} from Apple. The privacy guarantees of several of these APIs rely on \emph{differential privacy} (a.k.a. DP) \cite{DworkKMMN06,DworkMNS06}, which is a strong and robust notion of privacy that has in recent years gained significant popularity for data analytics and modeling tasks.

Two main DP variants have been studied in the context of supervised ML. So-called \emph{full} DP protects each individual training example (features and label) and has been extensively studied, e.g., \citet{abadi2016deep}. On the other hand, \emph{label} DP (e.g., \citet{chaudhuri2011sample,ghazi2021deep,malek2021antipodes}) only protects the label of each training example, and is thus a suitable option in settings where the adversary already has access to the features.

Label DP is a natural fit for the case where the features of the pCVR problem do not depend on cross-site information. However, a common setting, including that of the Protected Audience API on Chrome \cite{protected-audience-chrome} and Android \cite{protected-audience-android}, is where some features depend on cross-site information whereas the remaining features do not. An example is the remarketing use case where a feature could indicate whether the same user previously expressed interest in the advertised product (e.g., added it to their cart) but did not purchase it. Revealing a row of the database which has both contextual features (e.g., the publisher site, or the time of day the ad was served) with features derived on the advertiser (e.g., user presence on a particular remarketing list) could allow an attacker to track a user across sites.
In the Protected Audience API, these sensitive features are protected by multiple privacy mechanisms including feature-level randomized response. 

The focus of this work is to analyze this setting from DP perspective; we refer to it as \emph{DP model training with semi-sensitive features}. We formalize this setting, present an algorithm for training private ML models with semi-sensitive features, and evaluate it on real ad prediction datasets, showing that it compares favorably to natural baselines.

\section{Related Work}

We note that an evaluation of DP-SGD for ad prediction models in the full DP setting has been done by \citet{denison2022private}. We also point out that a stronger privacy guarantee than the DP with semi-sensitive features notion that we study, but that is still weaker than full DP, was recently introduced by \citet{krichene2023private}. In their ``DP with a public feature set'' setting, the adversary is assumed to know the set of values that a certain subset of features takes, while it ignores the correspondence between each example and its specific value.
In a concurrent work~\citep{shen2023classification}, an algorithm based on AdaBoost is proposed for a similar setting, referred to as learning with ``partially private features''. One key difference is that in their setting, the labels are considered public.

\section{DP Learning with Semi-Sensitive Features}\label{sec:prelim}
We consider the setting of supervised learning, where we assume an underlying (unknown) distribution $\cD$ over $\cX \times \cY$, where $\cX$ denotes the set of possible inputs and $\cY$ denotes the set of possible labels. In our work, we focus on the binary classification setting where $\cY = \{0, 1\}$. Our goal is to learn a predictor $F : \cX \to \R$ that maps the input space $\cX$ to $\R$ with the goal of minimizing the expected loss $\cL(F; \cD) := \Ex_{(x,y) \sim \cD} \ell(F(x), y)$, where $\ell(\cdot, \cdot)$ is a suitable loss function, e.g., the binary cross entropy loss.%

To capture the setting of {\em semi-sensitive features}, let $\cX = \cXns \times \cXs$, where $\cXns$ is the set of possible {\em non-sensitive feature values}, and $\cXs$ is the set of possible {\em sensitive feature values}.
We denote a dataset as $D = ((\xns_i, \xs_i, y_i))_{i \in [n]}$, where $\xns_i$ denotes the non-sensitive feature value, $\xs_i$ is the sensitive feature value and $y_i$ is the corresponding (sensitive) label. We use $x_i$ to denote $(\xns_i, \xs_i)$ for short.
Some problems that motivate the setting above are in ads modeling tasks, where the features can include non-sensitive features such as the browser class, publisher website, category of the mobile app etc.,  sensitive features such as how long ago and how many times a user showed interest in an advertised product etc.,
and sensitive labels such as whether the user converted on the ad.

We say that two datasets $D$, $D'$ are {\em adjacent}, denoted $D \sim D'$ if one dataset can be obtained from the other by changing the sensitive features and/or the label for a single example, namely replacing $(\xns_i, \xs_i, y_i)$ with $(\xns_i, \txs_i, \ty_i)$ for some $(\txs_i, \ty_i) \in \cXs \times \cY$.

\begin{definition}[DP;~\citet{DworkMNS06}]\label{def:differential_privacy}
For $\eps, \delta \geq 0$, a randomized mechanism $\cM$ satisfies \emph{$(\eps, \delta)$-$\DP$} if for all pairs $D, D'$ of adjacent datasets, and for all outcome events $E$, it holds that $\Pr[\cM(D) \in E] \leq e^\eps \cdot \Pr[\cM(D') \in E] + \delta$.
\end{definition}

\noindent For an extensive overview of DP, we refer the reader to the monograph of \citet{dwork2014algorithmic}. We use the following key properties of $\DP$.

\begin{proposition}[Composition]\label{prop:basic-composition}
If $\cM_1$ satisfies $(\eps_1, \delta_1)$-$\DP$, and $\cM_2$ satisfies $(\eps_2, \delta_2)$-$\DP$, then
the mechanism $\cM$ that on dataset $D$ returns $(\cM_1(D), \cM_2(D))$ satisfies $(\eps_1 + \eps_2, \delta_1 + \delta_2)$-$\DP$. Furthermore, this holds even in the adaptive case, when $\cM_2$ can use the output of $\cM_1$.
\end{proposition}

\begin{proposition}[Post-Processing]\label{prop:post-processing}
If $\cM$ satisfies $(\eps, \delta)$-$\DP$, then for all (randomized) algorithms $\cA$, it holds that $\cA(\cM(\cdot))$ satisfies $(\eps, \delta)$-$\DP$.
\end{proposition}

\paragraph{Randomized Response.}
Perhaps the simplest mechanism that satisfies DP, even predating its definition, is the {\em Randomized Response} mechanism. We state the mechanism in our context as releasing the known features along with the corresponding randomized label.

\begin{definition}[Randomized Response; \citet{warner1965randomized}]\label{def:rr}
For $\eps > 0$, the mechanism $\RR_{\eps}$ on dataset $D = ((\xns_i, \xs_i, y_i))_{i \in [n]}$ returns $((\xns_i, \hy_i))_{i \in [n]}$ where each $\hy_i$ is set to $y_i$ with prob. $\frac{e^\eps}{1 + e^\eps}$ and set to $1-y_i$ with prob. $\frac{1}{1 + e^\eps}$.
\end{definition}

\begin{proposition}\label{prop:rr}
$\RR_\eps$ satisfies $(\eps, 0)$-$\DP$.
\end{proposition}

\paragraph{\boldmath $\SGD$ and $\DPSGD$.}
Let $F_{\bw}$ be a parameterized model (e.g., a neural network) with trainable weights $\bw$, and $\{(x_1,y_1),\ldots,(x_B,y_B)\}$ be a random mini-batch of training examples. Let $L_i=\ell(F_\bw(x_i),y_i)$ be the loss on the $i$th example and let the average loss be $\bar{L} := \frac{1}{B}\sum_{i=1}^B L_i$.
Recall that standard training algorithms compute the \emph{average gradient} $\nabla_\bw\bar{L}$ and update $\bw$ with an optimizer such as SGD or Adam. Even though various optimizers could be used, we will refer to this class of (non-private) methods as $\SGD$.

$\DPSGD$~\citep{abadi2016deep} is widely used for DP training of deep neural networks, wherein the per-example gradients $\nabla_\bw L_i$ are computed, and then re-scaled to have an $\ell_2$-norm of at most $C$, as $\bg_i := \nabla_\bw L_i \cdot \min\{1, \frac{C}{\|\nabla_\bw L_i\|_2}\}$. Gaussian noise $\cN(\mathbf{0}, C^2\sigma^2 \bI)$ is then added to the average $\frac1B \sum_{i=1}^B \bg_i$ and subsequently passed to the optimizer.
As shown by \citet{abadi2016deep}, $\DPSGD$ satisfies $(\eps, \delta)$-$\DP$ where $\eps, \delta$ depend on $\sigma$, the batch size and number of training steps; this can be computed using privacy accounting.

\paragraph{User-level DP.}
In settings where a single-user may contribute more than one example to a dataset, it can be important to respect privacy at the user-level. To define this setting, it is assumed that $\xns$ contains a unique user identifier, and two datasets are said to be {\em user-level adjacent} if we can get one dataset from the other by changing the sensitive features and/or labels for one or more examples corresponding to a single user. A randomized mechanism is said to satisfy \emph{$(\eps, \delta)$-user-level-$\DP$} if the $(\eps,\delta)$-$\DP$ definition holds for user-level adjacent datasets $D, D'$. Any mechanism $\cM$ that satisfies $(\eps, \delta)$-$\DP$ can be transformed into a mechanism that satisfies user-level-$\DP$, by limiting the number of examples corresponding to any user. Namely, let $\cM^{(k)}$ be a mechanism that first processes the dataset by retaining at most $k$ examples corresponding to any single user (by applying an arbitrary rule that only considers the examples of that user), and then applies $\cM$ to it.

\begin{proposition}[Group Privacy; e.g., \citet{vadhan17complexity}]\label{prop:user-level-dp}
If $\cM$ satisfies $(\eps, \delta)$-$\DP$ then for all $k \ge 1$, any $\cM^{(k)}$ as defined above satisfies $(k \eps, \delta \frac{e^{k\eps - 1}}{e^\eps -1})$-user-level-$\DP$.
\end{proposition}

\section{Algorithms} \label{sec:our-algorithm}

We now describe the family of algorithms we use for DP training with semi-sensitive features. We use a parameterized model, such as a deep neural network that is parameterized by $\bw = (\wns, \ws, \wc)$ corresponding to a {\em non-sensitive} tower, a {\em sensitive} tower, and a {\em common} tower respectively. The architecture we use has the following high-level structure:
\[F_{\bw}(\xns, \xs) := f_{\wc}(g_{\wns}(\xns) \circ h_{\ws}(\xs))\,,\]
where $g_{\wns} : \cXns \to \R^{\dns}$, $h_{\ws} : \cXs \to \R^{\ds}$, $f_{\wc} : \R^{\dns + \ds} \to \R$ and $u \circ v$ denotes concatenation of the vectors $u$ and $v$.
We also consider a {\em truncated model} that uses the same parameters $\wns$ and $\wc$, but does not depend on $\ws$, by eliminating the dependence on $\xs$, defined as follows:
\[F_{\wns, \wc}(\xns) := f_{\wc}(g_{\wns}(\xns) \circ \mathbf{0})\,, \]
where $\mathbf{0} \in \R^{\ds}$.
For ease of notation, we use the following notation for the losses of each of these models:
\begin{align*}
    L(\bw; x, y) &~:=~ \ell(F_\bw(x), y)\\
    L(\wns, \wc; x, y) &~:=~ \ell(F_{\wns, \wc}(\xns), y)
\end{align*}
Without privacy constraints, we can consider training $F_{\bw}$ using $\SGD$ as described earlier.
Given a total privacy budget of $(\eps, \delta)$, we consider learning algorithms that execute two phases sequentially that satisfy $(\eps_1, 0)$-$\DP$ and $(\eps_2, \delta)$-$\DP$ respectively such that $\eps_1 + \eps_2 = \eps$ and hence by \Cref{prop:basic-composition}, the algorithm satisfies $(\eps, \delta)$-$\DP$. We refer to this algorithm as $\Hybrid$, and these phases are as follows:
\begin{description}[leftmargin=1mm]
\item [Label-DP Phase.] In this phase, we train the truncated model $F_{\wns, \wc}(\cdot)$ for one or more epochs of mini-batch $\SGD$ using noisy labels obtained by applying $\RR_{\eps_1}$, which returns $((\xns_i, \hy_i))_{i \in [n]}$, i.e., a dataset where the $\xs_i$'s are removed and the labels are randomized.
By \Cref{prop:post-processing}, this phase satisfies $(\eps_1, 0)$-$\DP$.

To remove the bias introduced by the noisy labels, we define $p := 1 / (1 + e^{-\eps_1})$ and modify the training loss as follows:
\begin{align*}
\frac{L(\wns, \wc; \xns_i , 1 - \hy_i) - p \sum_{y' \in \{0, 1\}} L(\wns, \wc; \xns_i, y')}{1 - 2p}
\end{align*}

\item [\boldmath $\DPSGD$ Phase.] In this phase, we train the entire model $F_{\bw}(\cdot)$ for one or more epochs of $\DPSGD$ on $F_{\bw}$. The noise parameter $\sigma$ is chosen appropriately so that this phase satisfies $(\eps_2, \delta)$-$\DP$; in our work, we do this accounting using R\'enyi DP~\cite{abadi2016deep,mironov17renyi}.
\end{description}

\section{Summary of Results}\label{sec:results}

\begin{figure}[h]
    \centering
    \includegraphics[width=0.95\columnwidth]{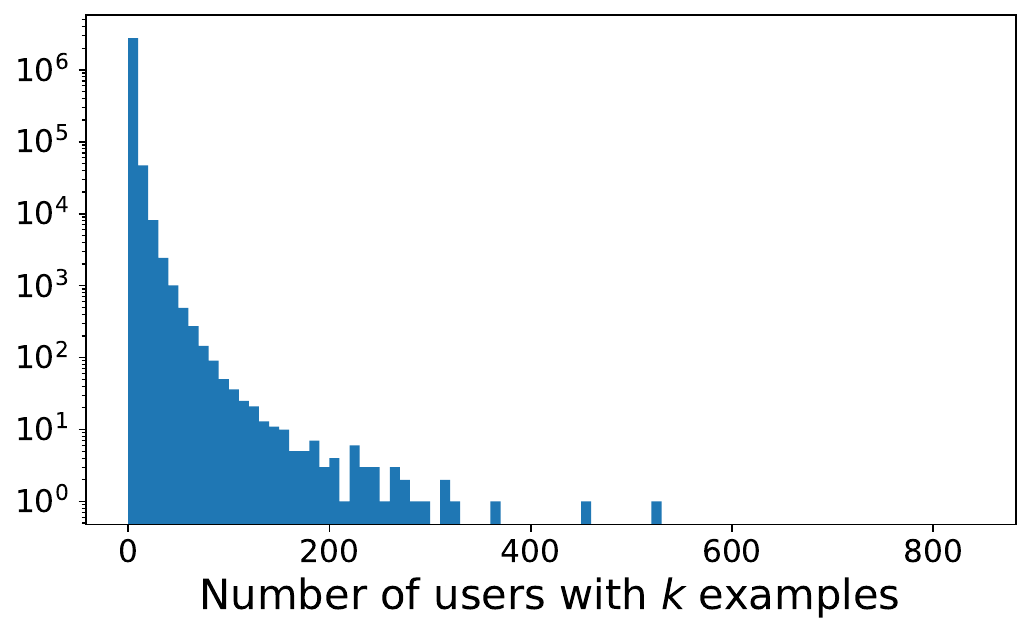}
    \caption{Number of examples per user in the Criteo Attribution dataset.}
    \label{fig:criteo-attribution-user-histogram}
\end{figure}

\begin{figure*}[h]
    \centering
    \begin{tabular}{cc}
    \includegraphics[width=0.95\columnwidth]{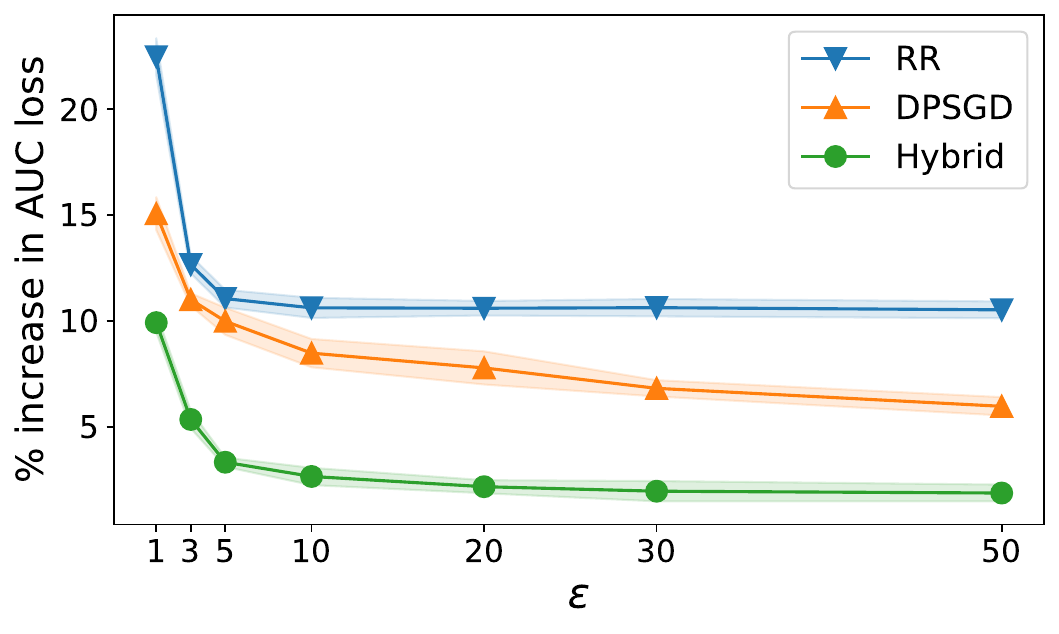} &
    \includegraphics[width=0.95\columnwidth] {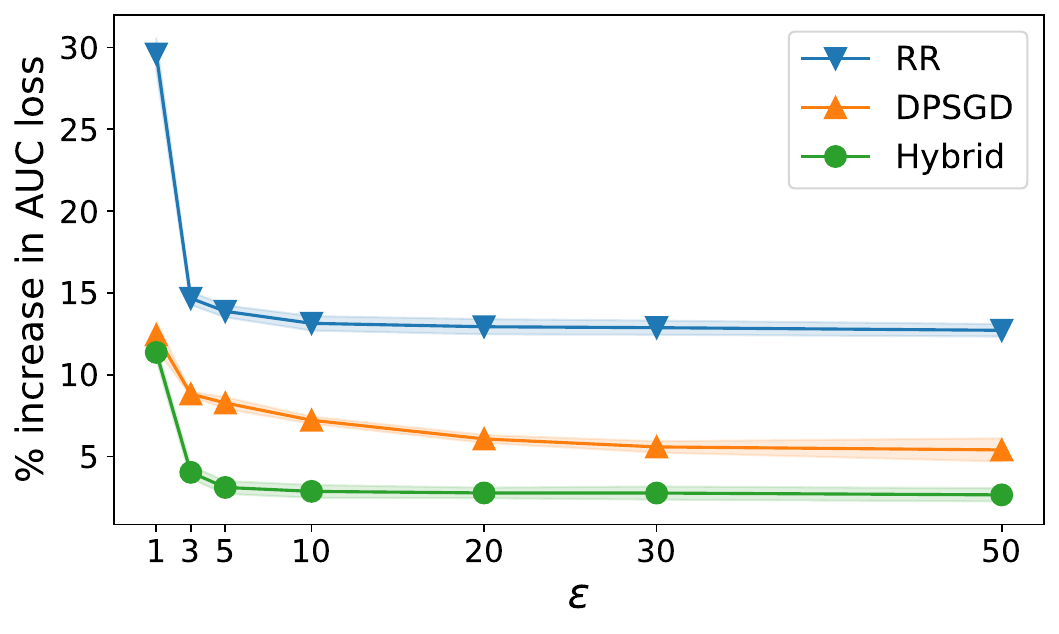} \\
    (i) & (ii)
    \end{tabular}
    \caption{Relative AUC loss (\%) of models trained under various privacy budget $\varepsilon$ on the Criteo Attribution dataset with (i) example-level DP and (ii) user-level DP, where for each $\varepsilon$, the lowest loss among the different example caps is plotted. }
    \label{fig:criteo-attribution-debiased}
\end{figure*}

\begin{figure*}[h]
    \centering
    \begin{tabular}{cc}
    \includegraphics[width=0.95\columnwidth]{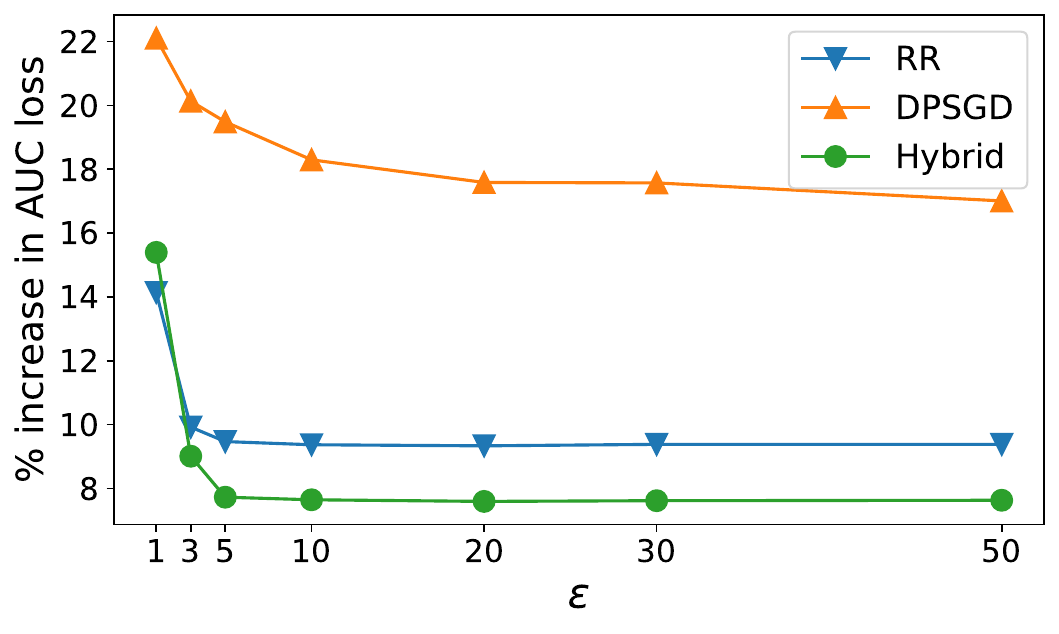} &
    \includegraphics[width=0.95\columnwidth]{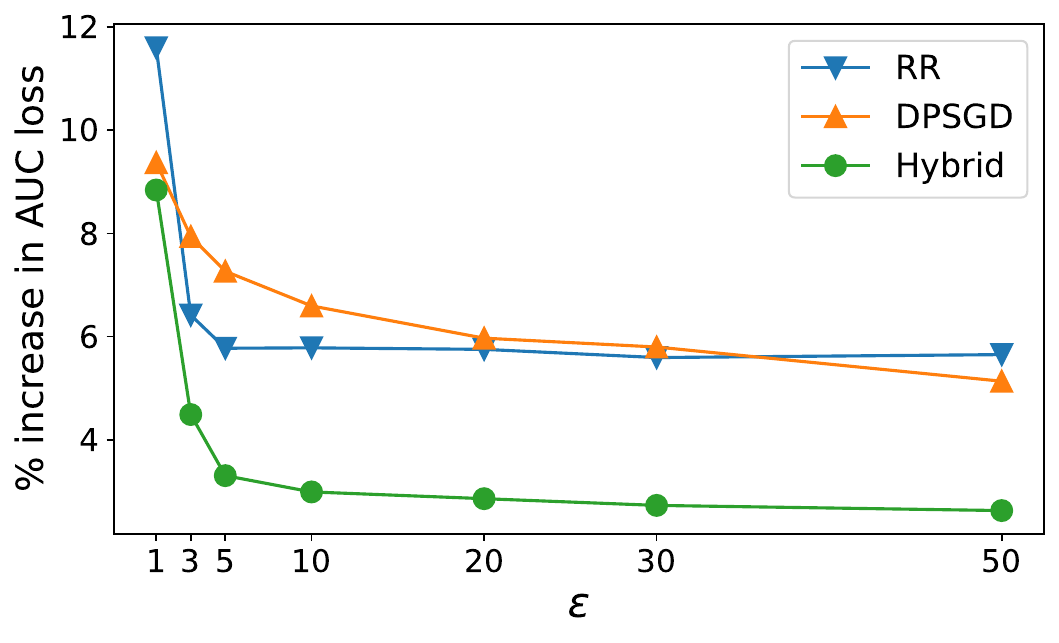} \\
    (i) & (ii) \\
    \end{tabular}
    \caption{Relative AUC loss (\%) of models trained under various privacy budget $\varepsilon$ on the (i)  Criteo pCTR dataset and (ii) proprietary pCVR dataset.}
    \label{fig:pctr-pcvr-debiased}
\end{figure*}

We consider two natural baselines: $\DPSGD$ (where all features are treated as sensitive) and $\RR$ on the truncated model (where the sensitive features are discarded and only the labels are protected). Note that both can be viewed as special cases of $\Hybrid$, where we use all the privacy budget in one of the phases: $\DPSGD$ corresponds to setting $\eps_1=0$ and $\eps_2 = \eps$ and $\RR$ corresponds to setting $\eps_1 = \eps$ and $\eps_2 = 0$.

The main algorithm $\Hybrid$ uses a different split between the two phases. Given a privacy budget of $(\eps, \delta)$, the precise split as $(\eps_1, 0)$ and $(\eps_2, \delta)$ between the two phases has a significant impact on the model's utility. We observed that as $\eps_1$ decreases beyond $3$, the $\RR$ utility deteriorates rapidly compared to $\DPSGD$. So when $\eps \approx 3$, we assign most of it to $\eps_1$. Beyond $\eps_1 = 4$, we observed that it is best to assign the remaining $\eps$ to $\DPSGD$, i.e, $\eps_2$. The precise split we use is as follows:
\[
\eps_1 ~:=~ \min\{0.6 \cdot \eps, 3\} \qquad \text{and} \qquad \eps_2 ~:=~ \eps - \eps_1
\]

We train binary classification models with binary cross entropy loss for all three datasets and report the test \emph{AUC loss-\%} with respect to the non-private baseline, i.e., $100 \cdot ((1-\AUC)-(1-\AUC_{\np})) /(1-\AUC_{\np})$, averaged over three runs with different random seeds. See the Appendix for more details about model architectures and training.

\paragraph{Criteo Attribution Modeling for Bidding Dataset.} This publicly available dataset \citep{diemert2017attribution} consists of ~16M impressions from 30 days of Criteo live traffic. We consider the task of predicting the Boolean \texttt{conversion} value (using last-touch attribution) using the impression attributes consisting of \texttt{campaign} and categorical features \texttt{cat[1-9]}. Since the precise interpretation of \texttt{cat[1-9]} is not available, we arbitrarily treat \texttt{cat1} and \texttt{cat2} as sensitive, and the remaining \texttt{cat[3-9]} and \texttt{campaign} as non-sensitive.

We randomly partition the dataset into 80\%/20\% train/test splits and report the results on the test split. We evaluate with both (example-level) $\DP$ as well as user-level-$\DP$ (using the transformation in \Cref{prop:user-level-dp}). For user-level-$\DP$, each user has varying numbers of examples, as illustrated in \Cref{fig:criteo-attribution-user-histogram}. We cap the number of examples per user, and sample examples with replacement up to the cap for each user. As a result, the size of the training datasets change with the cap. Hence instead of fixing the number of epochs for the $\DPSGD$ phase, we fix the number of training steps for each cap to be the same as for the largest cap trained with $1$ epoch.

For this dataset, the AUC loss of the non-privately trained model is $0.15879$. We plot the relative AUC loss with respect to this baseline in \Cref{fig:criteo-attribution-debiased}(i) for (example-level) $\DP$, and in \Cref{fig:criteo-attribution-debiased}(ii) for user-level-$\DP$.

\paragraph{\boldmath Criteo Display Ads $\mathsf{pCTR}$ Dataset.}

This publicly available dataset \cite{tien14criteokaggle} contains around ~40 million examples. The dataset has a labeled training set and an unlabeled test set. We only use the labeled training set and split it chronologically into a 80\%/10\%/10\% partition of train/validation/test sets. The reported metrics are on this labeled test split. Each example consists of 13 integer features \texttt{int-feature-[1-13]} and 26 categorical features \texttt{categorical-feature-[14-39]}. Since the precise interpretation of these features is not available, we arbitrarily consider all even-numbered features as sensitive and all odd-numbered features as non-sensitive.

For this dataset, the AUC loss of the non-privately trained model is $0.19422$. We plot the relative AUC loss with respect to this baseline in \Cref{fig:pctr-pcvr-debiased}(i). 

\paragraph{\boldmath Proprietary pCVR Dataset.} This is a proprietary dataset from a commercial mobile app store. Each example in this dataset corresponds to an ad click and the task is to predict whether a conversion takes place after the click. We plot the relative AUC loss with respect to this baseline in \Cref{fig:pctr-pcvr-debiased}(ii).

\medskip

For all datasets, we find that $\Hybrid$ improves over $\RR$ and $\DPSGD$ across a wide range of privacy budgets. We see that the gap increases for $\eps \geq 3$. We believe this is because the utility of the $\RR$ algorithm deteriorates for small $\eps$.

\section{Conclusion and Future Directions}

In this work, we studied training DP models with semi-sensitive features, and presented an algorithm that improves over two natural baselines on real ad modeling datasets.

A variant of DP training with semi-sensitive features that is interesting to explore is where the label is non-sensitive.
Another interesting direction is to determine the utility gap (if any) between ``DP with semi-sensitive features'' studied in this work and ``DP with a public feature set'' from \cite{krichene2023private}. Finally, it would be interesting to explore label DP primitives beyond $\RR$, e.g., \cite{ghazi2021deep,malek2021antipodes}, particularly ones that perform better for smaller $\eps$, for the label-DP phase of DP training with semi-sensitive features.

\bibliography{refs}

\appendix

\section{Training Details}

\paragraph{\boldmath Criteo Display Ads $\mathsf{pCTR}$ Dataset.}
We use a six layer neural network as our model. The first layer consists of feature transforms for each individual feature. For categorical features, we map the category into a dense feature vector using an embedding layer. We set the embedding dimension using the heuristic rule $\mathsf{int}[ 2 V^{0.25}]$, where $V$ is the number of unique tokens in each categorical feature. For int features, we apply a log transform. For our baseline model, all of these features are concatenated and fed into four fully connected layers, each containing $598$ hidden units and using a ReLU activation function. The final output is obtained using another fully connected layer which produces a scalar logit prediction. For our $\Hybrid$ model, we employ the same architecture and only consider the parameters in the sensitive categorical features' embedding layers as sensitive. There are a total of around $78$M trainable parameters in our model. 

We use the Yogi optimizer~\citep{zaheer2018adaptive} with a base learning rate of $0.01$ and a batch size of $1024$ for our baseline and for the $\RR$ phase of training. We use $\SGD$ with a base learning rate of $0.1$, momentum $0.9$, and batch size of $16384$ for the $\DPSGD$ phase of training. For both, we scale the base learning rate with a cosine decay~\citep{loshchilov2016sgdr}. For the baseline, we train for five epochs. We only tune the norm bound $C\in\{1, 10, 30, 50\}$, $\RR$ epochs $E_{\RR}\in\{1, 3, 50\}$, and $\DPSGD$ epochs $E_{\DPSGD}\in\{1, 3\}$, for each $\varepsilon$ separately. 

\begin{table}[]
    \centering
    \resizebox{\columnwidth}{!}{%
\begin{tabular}{cccc}
\toprule
$\varepsilon$ & $\RR$ & $\DPSGD$ & $\Hybrid$\\
\midrule
1 & $14.139\pm 0.001$ & $22.106\pm 0.002$ & $15.396\pm 0.007$\\
3 & $9.921\pm 0.000$ & $20.134\pm 0.001$ & $9.008\pm 0.002$\\
5 & $9.467\pm 0.000$ & $19.485\pm 0.001$ & $7.724\pm 0.001$\\
10 & $9.368\pm 0.000$ & $18.295\pm 0.002$ & $7.643\pm 0.002$\\
20 & $9.333\pm 0.000$ & $17.586\pm 0.001$ & $7.592\pm 0.002$\\
30 & $9.380\pm 0.000$ & $17.573\pm 0.001$ & $7.615\pm 0.001$\\
50 & $9.380\pm 0.001$ & $17.007\pm 0.001$ & $7.628\pm 0.002$\\
\bottomrule
\end{tabular}
}
    \caption{Relative AUC loss (\%) of models trained under various privacy budget $\varepsilon$ on the Criteo pCTR dataset.}
    \label{tab:criteo-pctr}
\end{table}

\paragraph{Criteo Attribution Modeling for Bidding Dataset.} 
The first layer of the model consists of feature transforms for each feature. For the categorical features, we map them into feature vectors using an embedding layer with embedding dimension $8$. These features are concatenated with the float features, and we use a layer to mask out sensitive features. These are fed into two layers of $128$ and $64$ hidden units. The final output is obtained using a fully-connected layer which produces a scalar logit prediction. There are a total of around $0.5$M trainable parameters in the model.

For the training, we use the RMSprop optimizer with a learning rate of $0.0001$ and a batch size of $8192$ for our baseline and for the $\RR$ training phase. We use the Adam optimizer with a learning rate of $0.0001$ and a batch size of $16384$ for the $\DPSGD$ training phase. For both, we scale the base learning rate with a cosine decay. For the baseline, we train for $50$ epochs. We tune the norm bound $C\in\{1, 5, 10, 50, 100\}$, $\RR$ epochs $E_{\RR}\in\{1, 3, 5, 10, 50, 200\}$, example caps $k\in\{1, 2, 5, 10\}$, for each $\eps$ separately. For the $\Hybrid$ training, we set the example cap to be $1$ for the $\RR$ phase and we vary it as $k\in\{1, 2, 5, 10\}$ for the $\DPSGD$ phase., For (example-level) $\DP$, we tune the $\DPSGD$ epochs $E_{\DPSGD}\in\{1, 3\}$. For user-level-$\DP$, we sample examples independently with replacement up to the example cap for each user, and we fix the number of training steps for each cap to be the same as for the largest cap trained with one epoch. The relative AUC losses with respect to the non-private baseline are in \Cref{fig:criteo-attribution-debiased}(i), \Cref{tab:criteo-attribution} for (example-level) $\DP$ and in \Cref{fig:criteo-attribution-debiased}(ii), \Cref{tab:criteo-attribution-user} for user-level-$\DP$.

\begin{table}[]
    \centering
\begin{tabular}{cccc}
\toprule
$\varepsilon$ & $\RR$ & $\DPSGD$ & $\Hybrid$\\
\midrule
1 & $22.481\pm 0.879$ & $15.068\pm 0.779$ & $9.919\pm 0.448$\\
3 & $12.658\pm 0.418$ & $11.007\pm 0.299$ & $5.343\pm 0.418$\\
5 & $11.054\pm 0.418$ & $9.983\pm 0.642$ & $3.322\pm 0.217$\\
10 & $10.616\pm 0.477$ & $8.477\pm 0.669$ & $2.650\pm 0.409$\\
20 & $10.599\pm 0.344$ & $7.778\pm 0.784$ & $2.166\pm 0.308$\\
30 & $10.625\pm 0.412$ & $6.814\pm 0.385$ & $1.953\pm 0.480$\\
50 & $10.523\pm 0.398$ & $5.965\pm 0.434$ & $1.867\pm 0.398$\\
\bottomrule
\end{tabular}
    \caption{Relative AUC loss (\%) of models trained under various privacy budget $\varepsilon$ on the Criteo Attribution dataset. }
    \label{tab:criteo-attribution}
\end{table}

\begin{table}[]
    \centering
    \resizebox{\columnwidth}{!}{%
\begin{tabular}{ccccc}
\toprule
$\varepsilon$ & Example cap & $\RR$ & $\DPSGD$ & $\Hybrid$ \\
\midrule
1 & 1 & $33.510\pm 0.678$ & $12.433\pm 0.830$ & $11.369\pm 0.446$\\
1 & 2 & $29.602\pm 0.967$ & $13.882\pm 1.199$ & $11.959\pm 0.536$\\
1 & 5 & $46.503\pm 0.804$ & $19.259\pm 0.668$ & $13.886\pm 0.377$\\
1 & 10 & $82.281\pm 3.847$ & $21.776\pm 0.326$ & $15.197\pm 0.533$\\
\midrule
3 & 1 & $14.675\pm 0.400$ & $8.827\pm 0.129$ & $4.040\pm 0.394$\\
3 & 2 & $19.346\pm 0.079$ & $10.097\pm 0.250$ & $4.259\pm 0.413$\\
3 & 5 & $22.287\pm 0.317$ & $13.312\pm 0.382$ & $4.582\pm 0.434$\\
3 & 10 & $37.793\pm 2.093$ & $14.913\pm 1.048$ & $5.220\pm 0.433$\\
\midrule
5 & 1 & $13.878\pm 0.364$ & $8.280\pm 0.355$ & $3.118\pm 0.390$\\
5 & 2 & $14.894\pm 0.325$ & $8.957\pm 0.533$ & $3.174\pm 0.392$\\
5 & 5 & $17.145\pm 0.326$ & $11.664\pm 0.285$ & $3.482\pm 0.316$\\
5 & 10 & $23.605\pm 0.644$ & $13.859\pm 0.975$ & $4.156\pm 0.371$\\
\midrule
10 & 1 & $13.299\pm 0.466$ & $7.225\pm 0.222$ & $2.878\pm 0.391$\\
10 & 2 & $13.145\pm 0.442$ & $8.012\pm 0.515$ & $2.959\pm 0.461$\\
10 & 5 & $14.438\pm 0.353$ & $10.250\pm 0.316$ & $3.107\pm 0.377$\\
10 & 10 & $15.735\pm 0.490$ & $11.618\pm 1.220$ & $3.341\pm 0.369$\\
\midrule
20 & 1 & $13.322\pm 0.407$ & $6.083\pm 0.243$ & $2.782\pm 0.325$\\
20 & 2 & $12.936\pm 0.465$ & $7.484\pm 0.417$ & $2.825\pm 0.429$\\
20 & 5 & $13.235\pm 0.380$ & $8.843\pm 0.456$ & $2.974\pm 0.448$\\
20 & 10 & $13.515\pm 0.425$ & $10.601\pm 0.269$ & $3.087\pm 0.367$\\
\midrule
30 & 1 & $13.394\pm 0.432$ & $5.594\pm 0.344$ & $2.812\pm 0.315$\\
30 & 2 & $12.971\pm 0.476$ & $7.152\pm 0.309$ & $2.776\pm 0.402$\\
30 & 5 & $12.880\pm 0.433$ & $9.002\pm 0.374$ & $2.929\pm 0.388$\\
30 & 10 & $13.076\pm 0.603$ & $9.767\pm 0.165$ & $3.106\pm 0.367$\\
\midrule
50 & 1 & $13.293\pm 0.396$ & $5.411\pm 0.710$ & $2.663\pm 0.402$\\
50 & 2 & $12.956\pm 0.372$ & $6.598\pm 0.349$ & $2.669\pm 0.389$\\
50 & 5 & $12.893\pm 0.374$ & $7.993\pm 0.154$ & $2.926\pm 0.348$\\
50 & 10 & $12.718\pm 0.381$ & $9.872\pm 1.159$ & $3.085\pm 0.403$\\
\bottomrule
\end{tabular}
}
    \caption{Relative AUC loss (\%) of models trained under various privacy budget $\varepsilon$ on the Criteo Attribution dataset. }
    \label{tab:criteo-attribution-user}
\end{table}

\paragraph{Hyperparameter Tuning}
We do not perform extensive hyperparameter tuning for these experiments and rely on past experience and intuition to set hyperparameters unless otherwise stated. We use existing baselines to set our baseline hyperparameters. For the $\RR$ phase, we find that these same hyperparameters work well. For the $\DPSGD$ phase, we rely on the work of \citet{denison2022private} for setting hyperparameters for the Criteo pCTR dataset. For Criteo Attribution, we tune the hyperparameters with different optimizers, learning rates and batch sizes. We did not do any hyperparameter tuning for the pCVR dataset but relied on past experience with the dataset.

\begin{table}[]
    \centering
    \resizebox{\columnwidth}{!}{%
\begin{tabular}{cccc}
\toprule
$\varepsilon$ & $\RR$ & $\DPSGD$ & $\Hybrid$\\
\midrule
1 & $11.595\pm 0.009$ & $9.371\pm 0.003$ & $8.840\pm 0.001$\\
3 & $6.418\pm 0.001$ & $7.940\pm 0.001$ & $4.491\pm 0.003$\\
5 & $5.775\pm 0.005$ & $7.266\pm 0.001$ & $3.308\pm 0.000$\\
10 & $5.783\pm 0.003$ & $6.595\pm 0.001$ & $2.992\pm 0.000$\\
20 & $5.754\pm 0.001$ & $5.971\pm 0.002$ & $2.862\pm 0.001$\\
30 & $5.595\pm 0.001$ & $5.799\pm 0.002$ & $2.732\pm 0.000$\\
50 & $5.652\pm 0.004$ & $5.137\pm 0.001$ & $2.631\pm 0.001$\\
\bottomrule
\end{tabular}
}
    \caption{Relative AUC loss (\%) of models trained under various privacy budget $\varepsilon$ on the proprietary pCVR dataset.}
    \label{tab:appads-debiased}
\end{table}

\section{Large Epoch Training for $\DPSGD$}
\begin{figure}
    \centering
    \includegraphics[width=\columnwidth]{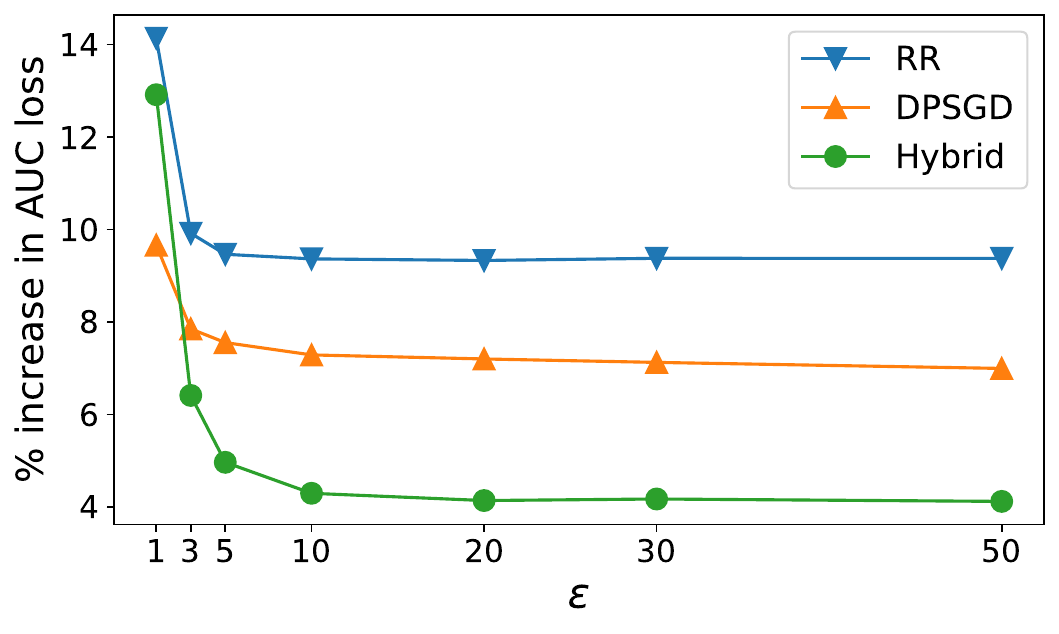}
    \caption{Relative AUC loss (\%) of models trained under various privacy budget $\varepsilon$ on the Criteo pCTR dataset with 64 epochs for $\DPSGD$ phase.}
    \label{fig:criteo-large-debiased}
\end{figure}

Due to limited compute and the high cost of $\DPSGD$ training, we only evaluated models where the $\DPSGD$ phase of training runs for at most three epochs. We find that for large datasets such our proprietary pCVR dataset, we are not practically able to run more epochs of training. To directly compare our algorithm with that of \citet{denison2022private} which uses $64$ epochs for their $\DPSGD$ training on the Criteo pCTR dataset, we also repeated the experiments on this dataset with $\DPSGD$ training run for $64$ epochs. All other hyperparameters remain the same. We find that this allows $\DPSGD$ to perform much better but our algorithm $\Hybrid$ still outperforms it for $\eps \geq 3$. We plot the relative AUC loss with respect to the baseline in \Cref{fig:criteo-large-debiased} and add the AUC loss values in \Cref{tab:criteo-pctr-large}.

\begin{table}[]
    \centering
    \resizebox{\columnwidth}{!}{%
\begin{tabular}{cccc}
\toprule
$\varepsilon$ & $\RR$ & $\DPSGD$ & $\Hybrid$\\
\midrule
1 & $14.139\pm 0.001$ & $9.666\pm 0.002$ & $12.918\pm 0.004$\\
3 & $9.921\pm 0.000$ & $7.854\pm 0.002$ & $6.415\pm 0.002$\\
5 & $9.467\pm 0.000$ & $7.557\pm 0.002$ & $4.968\pm 0.000$\\
10 & $9.368\pm 0.000$ & $7.291\pm 0.001$ & $4.298\pm 0.001$\\
20 & $9.333\pm 0.000$ & $7.203\pm 0.001$ & $4.142\pm 0.002$\\
30 & $9.380\pm 0.000$ & $7.129\pm 0.001$ & $4.175\pm 0.001$\\
50 & $9.380\pm 0.001$ & $6.999\pm 0.001$ & $4.123\pm 0.001$\\
\bottomrule
\end{tabular}
}
    \caption{Relative AUC loss (\%) of models trained under various privacy budget $\varepsilon$ on the Criteo pCTR dataset with 64 epochs for $\DPSGD$ phase.}
    \label{tab:criteo-pctr-large}
\end{table}

\end{document}